\documentclass[runningheads,a4paper]{llncs}

\usepackage{amssymb}
\usepackage[utf8]{inputenc}
\usepackage{graphicx}
\usepackage{times}
\usepackage{latexsym}
\usepackage{todonotes}
\usepackage{dcolumn}
\usepackage{graphics}
\usepackage{url}
\usepackage{longtable}
\usepackage{supertabular}
\usepackage{pdflscape}
\usepackage{pifont}

\usepackage{rotating}
\usepackage{tabularx}
\usepackage{lscape}
\usepackage{multirow}
\usepackage{multicol}
\usepackage{float}
\usepackage{footnote}
\usepackage{verbatim}
\usepackage{hyperref}

\usepackage{url}
\usepackage{mathtools}
\usepackage{scrextend}
\usepackage{xspace}
\usepackage{listingsutf8}
\usepackage{color}
\usepackage{booktabs}
\usepackage{tikz}
\usepackage{lipsum}
\usepackage[ruled,vlined]{algorithm2e}
\usepackage{mdframed}
\usepackage{wrapfig}
\usepackage[nolist]{acronym}
\usepackage{adjustbox}
\usepackage{array}
\usepackage{cite}

\newcolumntype{R}[2]{%
    >{\adjustbox{angle=#1,lap=\width-(#2)}\bgroup}%
    l%
    <{\egroup}%
}

\newcommand{\tablefont}[1]{\fontsize{3mm}{3.2mm}\selectfont}

\usepackage{listings}
\definecolor{olivegreen}{rgb}{0.2,0.8,0.5}
\definecolor{grey}{rgb}{0.5,0.5,0.5}

\lstdefinelanguage{ttl}{
  basicstyle=\footnotesize \ttfamily,
  sensitive=true,
  morecomment=[l][\color{grey}]{@},
  morecomment=[l][\color{olivegreen}]{\#},
  morestring=[b][\color{blue}]\",
}

\lstdefinestyle{rdf}{numberblanklines=false, morekeywords={},
backgroundcolor=\color{backcolour},   
    commentstyle=\color{codegreen},
    keywordstyle=\color{magenta},
    stringstyle=\color{codeblue},
    basicstyle=\footnotesize,
    breakatwhitespace=false,         
    breaklines=true,                 
    captionpos=b,                    
    keepspaces=true,                 
    numbers=none,                    
    numbersep=5pt,                  
    showspaces=false,                
    showstringspaces=false,
    showtabs=false,                  
    tabsize=1
}

\lstdefinelanguage{SPARQL}{%
   morekeywords=[1]{CONSTRUCT,WHERE,SELECT},
   morekeywords=[2]{AND,FILTER,UNION,OPT,OPTIONAL,MINUS,ORDER,GROUP,BY,DESC,OFFSET,LIMIT},%
   morekeywords=[3]{sameTerm,isBLANK,isLITERAL,isIRI,BOUND,DISTINCT},
   morekeywords=[4]{rdf,rdfs,owl,dbo,res,xsd},
   morekeywords=[5]{>},
   morestring=[b]",%
   alsodigit={-},%
}[keywords,strings]

\begin{acronym}[UML]
	\acro{AOS}{Agricultural Ontology Services}
	\acro{AGRIS}{Agricultural Science and Technology}
	\acro{API}{Application Programming Interface}
	\acro{A2KB}{Annotation to Knowledge Base}
	\acro{BPSO}{Binary Particle-Swarm Optimization}
	\acro{BPMLOD}{Best Practices for Multilingual Linked Open Data}
	\acro{BFS}{Breadth-First-Search}
	\acro{CBD}{Concise Bounded Description}
	\acro{COG}{Content Oriented Guidelines}
	\acro{CSV}{Comma-Separated Values}
	\acro{CCR}{cross-document co-reference}
	\acro{DPSO}{Deterministic Particle-Swarm Optimization}
	\acro{DALY}{Disability Adjusted Life Year}
	\acro{D2KB}{Disambiguation to Knowledge Base}

	\acro{ER}{Entity Resolution}
	\acro{EM}{Expectation Maximization}
	\acro{EL}{Entity Linking}
	\acro{FAO}{Food and Agriculture Organization of the United Nations}
	\acro{GIS}{Geographic Information Systems}
	\acro{GHO}{Global Health Observatory}
	\acro{HDI}{Human Development Index}
	\acro{ICT}{Information and communication technologies}
    \acro{KB}{Knowledge Base}
	\acro{LR}  {Language Resource}
	\acro{LD}  {Linked Data}
	\acro{LLOD}  {Linguistic Linked Open Data}
	\acro{LIMES}{LInk discovery framework for MEtric Spaces}
	\acro{LS}  {Link Specifications}
	\acro{LDIF}{Linked Data Integration Framework}
	\acro{LGD} {LinkedGeoData}
	\acro{LOD} {Linked Open Data}
	\acro{MSE}{Mean Squared Error}
	\acro{MWE}{Multiword Expressions}
	\acro{NIF}{NLP Interchange Format}
	\acro{NIF4OGGD}{NLP Interchange Format for Open German Governmental Data}
	\acro{NLP}{Natural Language Processing}
	\acro{NER}{Named Entity Recognition}
	\acro{NED}{Named Entity Disambiguation}
	\acro{NEL}{Named Entity Linking}
	\acro{NN}{Neural Network}
	\acro{OSM}{OpenStreetMap}
	\acro{OWL}{Web Ontology Language}
	\acro{PFM}{Pseudo-F-Measures}
	\acro{PSO}{Particle-Swarm Optimization}
	\acro{QA}{Question Answering}
	\acro{RDF}{Resource Description Framework}
	\acro{SKOS}{Simple Knowledge Organization System}
	\acro{SPARQL}{SPARQL Protocol and RDF Query Language}
	\acro{SRL}{Statistical Relational Learning}
	\acro{SF}{surface form}
	\acro{UML}{Unified Modeling Language}
	\acro{WHO}{World Health Organization}
	\acro{WKT}{Well-Known Text}
	\acro{W3C}{World Wide Web Consortium}
	\acro{YPLL}{Years of Potential Life Lost}
\end{acronym}  

\begin{document}

\mainmatter  

\title{Entity Linking in 40 Languages using MAG}

\titlerunning{Linking 40 languages}

\author{
Diego Moussallem\inst{1,2} \and
Ricardo Usbeck\inst{2}\orcidID{0000-0002-0191-7211} 
\and
Michael Röder\inst{2} \and \\
Axel-Cyrille Ngonga Ngomo\inst{1,2} }


\institute{AKSW Research Group, University of Leipzig, Germany\and 
Data Science Department, Paderborn University, Germany\\
\email{lastname@informatik.uni-leipzig.de} \\
}

\authorrunning{Moussallem et al.}

\toctitle{MAG}
\tocauthor{Authors' Instructions}
\maketitle

\begin{abstract}
A plethora of  Entity Linking (EL) approaches has recently been developed. While many claim to be multilingual, the MAG (Multilingual AGDISTIS) approach has been shown recently to outperform the state of the art in multilingual EL on 7 languages.  
With this demo, we extend MAG to support EL in 40 different languages, including especially low-resources languages such as Ukrainian, Greek, Hungarian, Croatian, Portuguese, Japanese and Korean. Our demo relies on online web services which allow for an easy access to our entity linking approaches and can disambiguate against DBpedia and Wikidata. During the demo, we will show how to use MAG by means of POST requests as well as using its user-friendly web interface.
All data used in the demo is available at \url{https://hobbitdata.informatik.uni-leipzig.de/agdistis/}
\end{abstract}

\section{Introduction}
\label{sec:introduction}

A recent survey by IBM\footnote{\url{https://tinyurl.com/ibm2017stats}} suggests that more than 2.5 quintillion bytes of data are produced on the Web every day. \ac{EL}, also known as \ac{NED}, is one of the most important \ac{NLP} techniques for extracting knowledge automatically from this huge amount of data. The goal of an \ac{EL} approach is as follows: Given a piece of text, a reference knowledge base $K$ and a set of entity mentions in that text, map each entity mention to the corresponding resource in $K$ \cite{moussallem2017mag}. A large number of challenges has to be addressed while performing a disambiguation. For instance, a given resource can be referred to using different labels due to phenomena such as synonymy, acronyms or typos. For example, \texttt{New York City}, \texttt{NY} and \texttt{Big Apple} are all labels for the same entity. Also, multiple entities can share the same name due to homonymy and ambiguity. For example, both the state and the city of Rio de Janeiro are called \texttt{Rio de Janeiro}. 

Despite the complexity of the task, \ac{EL} approaches have recently achieved increasingly better results by relying on trained machine learning models~\cite{roder2017gerbil}. A portion of these approaches claim to be multilingual and most of them rely on models which are trained on English corpora with cross-lingual dictionaries. 
However, MAG (Multilingual AGDISTIS)~\cite{moussallem2017mag} showed that the underlying models being trained on English corpora make them prone to failure when migrated to a different language. Additionally, these approaches hardly make their models or data available on more than three languages~\cite{roder2017gerbil}. The new version of MAG (which is the quintessence of this demo) provides support for 40 different languages using sophisticated indices\footnote{The quality of indices is directly related to how much information is provided by Wikipedia and DBpedia}. 
For the sake of server space, we deployed MAG-based web services for 9 languages and offer the other 31 languages for download. Additionally, we provide an English index using Wikidata to show the knowledge-base agnosticism of MAG. During the demo, we will show how to use the web services as well as MAG's user interface.

\section{MAG Entity Linking System}
\label{sec:approach}
  
MAG's \ac{EL} process comprises two phases, namely an offline and an online phase.
The sub-indices (which are generated during the offline phase) consist of surface forms, person names, rare references, acronyms and context information. 
During the online phase, the \ac{EL} is carried out in two steps: 1) candidate generation and 2) disambiguation. 
The goal of the candidate generation step is to retrieve a tractable number of candidates for each mention. 
These candidates are later inserted into the disambiguation graph, which is used to determine the mapping between entities and mentions. 
MAG implements two graph-based algorithms to disambiguate entities, i.e., PageRank and HITS. 
Independently of the chosen graph algorithm, the highest candidate score among the set of candidates is chosen as correct disambiguation for a given mention~\cite{moussallem2017mag}. 

\section{Demonstration}
\label{sec:demo}

\begin{figure}[htb]
\centering
\includegraphics[scale=0.30]{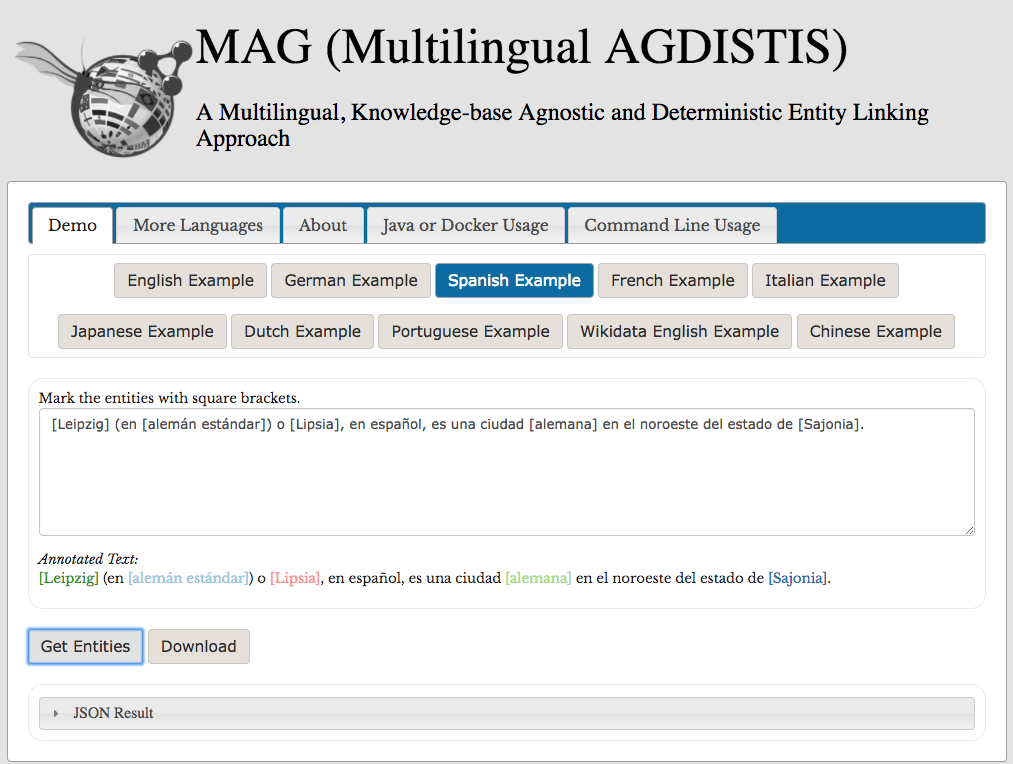}
\caption{A screenshot of MAG's web-based demo working on Spanish. }
\label{fig:spanish}
\end{figure}

Our demonstration will show the capabilities of MAG for different languages. We provide a graphical, web-based user interface (GUI).
In addition, users can choose to use the REST interface or a Java snippet. 
For research purposes, MAG can be downloaded and deployed via Maven or Docker. ~\autoref{fig:spanish} illustrates an example of MAG working on Spanish. The online demo can be accessed via \url{http://agdistis.aksw.org/mag-demo} and its code can be downloaded from \url{https://github.com/dice-group/AGDISTIS_DEMO/tree/v2}.

We have set up a web service interface for each language version. 
Each of these interfaces understands two mandatory parameters: (1) \texttt{text} and (2) \texttt{type}.  

\begin{enumerate}
    \item  \texttt{text} accepts an UTF-8 and URL encoded string with entities annotated with XML-tag \texttt{<entity>}. It is also capable of recognizing NIF\cite{NIF} or txt files.
    
    \item \texttt{type} accepts two different values. First, \texttt{'agdistis'} to disambiguate the mentions using the graph-based algorithms, but also \texttt{'candidates'} which list all possible entities for a given mention through the depth-candidate selection of MAG. 
\end{enumerate}

\noindent \textbf{Other parameters.} The user can also define more parameters to fine-tune the disambiguation. 
These parameters have to be set up within the properties file\footnote{\url{https://tinyurl.com/agdistis-properties}} or via environment variables while deploying it locally.
Below, we describe all the parameters.

\begin{itemize}
    \item \textbf{Popularity} - The user can set it as \texttt{popularity=false} or \texttt{popularity=true}. 
    It allows MAG to use either the Page Rank or the frequency of a candidate to sort while candidate retrieval. 
    
    \item \textbf{Graph-based algorithm} - The user can choose which graph-based algorithm to use for disambiguating among the candidates per mentions. The current implementation offers HITS and PageRank as algorithms, \texttt{algorithm=hits} or \texttt{algorithm =pagerank}.
    
    \item \textbf{Search by Context} - This boolean parameter provides a search of candidates using a context index~\cite{moussallem2017mag}. 
    
    \item \textbf{Acronyms} -  This parameter enables a search by acronyms. In this case, MAG uses an additional index to filter the acronyms by expanding their labels and assigns them a high probability. For example, PSG equals Paris Saint-Germain. The parameter is \texttt{acronym=false} or \texttt{acronym=true}.
    
    \item \textbf{Common Entities} - This boolean option supports finding common entities , in case,  users desire to find more than ORGANIZATIONs, PLACEs and PERSONs as entity type.  
    
    \item \textbf{Ngram Distance} - This integer parameter chooses the ngram distance between words, e.g., bigram, trigram and so on.

    \item \textbf{Depth} -  This parameter numerically defines how deep the exploration of a semantic disambiguation graph must go.

    \item \textbf{Heuristic Expansion} - This boolean parameter defines whether a simple co-occur\-ren\-ce resolution is done or not. For instance, if Barack and Barack Obama are in the same text then Barack is expanded to Barack Obama.
\end{itemize}

\begin{figure}[htb]
\centering
\includegraphics[scale=0.30]{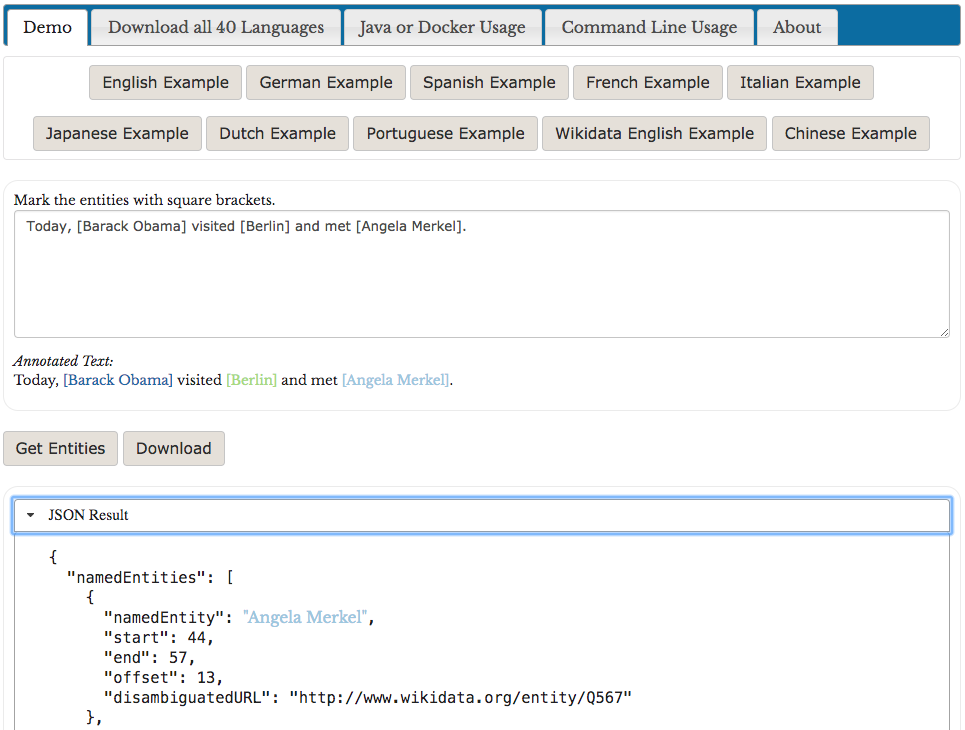}
\caption{MAG working on Wikidata as Knowledge base.}
\label{fig:wikidata}
\end{figure}

\noindent \textbf{Knowledge-base Agnosticism.} \autoref{fig:wikidata} shows a screen capture of our demo for disambiguating mentions using Wikidata. We also provide a web service to allow further investigation. In addition, MAG is used in a domain specific problem using a music \ac{KB}~\cite{oramasmel}.

\section{Evaluation of the user interface}

\begin{figure}[htb]
	\centering
	\includegraphics[width=0.6\columnwidth]{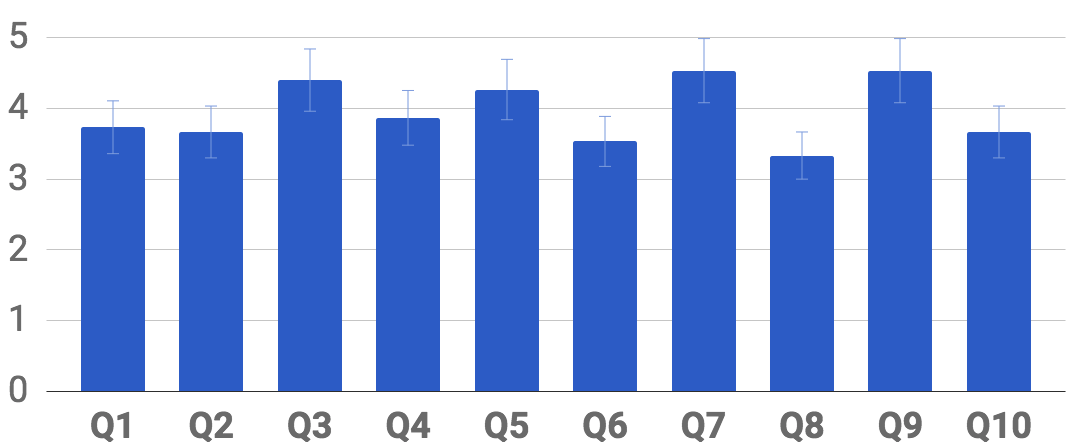}
	\caption{Standard Usability Score results. The vertical bars show the standard deviation}	
	\label{fig:sus}
\end{figure}

We performed a system usability study (SUS)\footnote{\url{http://www.measuringu.com/sus.php}}\footnote{\url{https://goo.gl/forms/01kpxBf24pjbsWUV2}} to validate the design of our user interface.
15 users -  with a good or no knowledge of Semantic Web, \ac{EL} or knowledge extraction - selected randomly from all departments at Leipzig University answered our survey. We achieved a SUS-Score of 86.3. 
This score assigns the mark $S$ to the current interface of MAG and places it into the \emph{category of the 10\% interfaces}, meaning that users of the interface are likely to recommend it to a friend.
~\autoref{fig:sus} shows the average voting per question and its standard deviation.

\section{Summary}
 \label{sec:conclusion}

In this demo, we will present MAG, a \ac{KB}-agnostic and deterministic approach for multilingual \ac{EL} on 40 different languages contained in DBpedia. Currently, MAG is used in diverse projects\footnote{For example, \url{http://diesel-project.eu/}, \url{https://qamel.eu/} or \url{https://www.limbo-project.org/}} and has been used largely by the Semantic Web community.
We also provide a demo/web-service using Wikidata for supporting an investigation of the graphs structures behind DBpedia and Wikidata pertaining to Information Extraction tasks\cite{geiss2017neckar,farber2015comparative}. 
The indexes we provided will be used in future work to investigate the \ac{EL} problem in low-resource languages. Our next step will hence be to to evaluate \ac{EL} on all 40 languages presented in this demo.


\textbf{Acknowledgements} This work has been supported by the BMVI projects LIMBO (project no. 19F2029C) and OPAL (project no. 19F20284) as well as by the German Federal Ministry of Education and Research (BMBF) within ’KMU-innovativ: Forschung für die zivile Sicherheit’ in particular ’Forschung für die zivile Sicherheit’ and the project SOLIDE (no. 13N14456). This work has also been supported by the Brazilian National Council for Scientific and Technological Development (CNPq) (no. 206971/2014-1). The authors gratefully acknowledge financial support from the German Federal Ministry of Education and Research within Eurostars, a joint programme of EUREKA and the European Community under the project E! 9367 DIESEL and E! 9725 QAMEL.

\bibliographystyle{abbrv}
\bibliography{agdistis_ext}

\end{document}